\documentclass[sigconf]{acmart}

\usepackage[utf8]{inputenc} 
\usepackage[T1]{fontenc}    
\usepackage{hyperref}       
\usepackage{url}            
\usepackage{booktabs}       
\usepackage{amsfonts}       
\usepackage{nicefrac}       
\usepackage{microtype}      

\usepackage{wrapfig}
\usepackage{times}
\usepackage{soul}

\usepackage{subcaption}
\usepackage{amsmath}
\usepackage{amsthm}

\usepackage{tabularray}
\usepackage{algorithmicx}  
\usepackage[linesnumbered,ruled,vlined]{algorithm2e}
\usepackage{algpseudocode}  
\usepackage{amsmath}  
\usepackage{multirow}

\usepackage{newfloat}
\usepackage{listings} 
\newtheorem{theorem}{Theorem}
\AtBeginDocument{%
  }

\copyrightyear{2026}
\acmYear{2026}
\setcopyright{cc}
\setcctype{by-nc-nd}
\acmConference[KDD 2026] {Proceedings of the 32nd ACM SIGKDD Conference on Knowledge Discovery and Data Mining V.1}{August 9--13, 2025}{Jeju Island, Republic of Korea.}
\acmBooktitle{Proceedings of the 32nd ACM SIGKDD Conference on Knowledge Discovery and Data Mining V.1 (KDD 2026), August 9--13, 2025, Jeju Island, Republic of Korea}
\acmISBN{979-8-4007-2258-5/2026/08}
\acmDOI{10.1145/3770854.3780308}

\begin{document}

\title{CFLight: Enhancing Safety with Traffic Signal Control through Counterfactual Learning}


\author{Mingyuan Li}
\email{henryli_i@bupt.edu.cn}
\affiliation{%
  \institution{Lanzhou University;\\ Beijing University of Posts and Telecommunications,}
  \city{Lanzhou}
  \country{China}
}

\author{Chunyu Liu}
\email{chunyuliu@bupt.edu.cn}
\affiliation{%
  \institution{Beijing University of Posts and Telecommunications,}
  \city{Beijing}
  \country{China}
}

\author{Zhuojun Li}
\email{zhuojinli@bupt.edu.cn}
\affiliation{%
  \institution{Beijing University of Posts and Telecommunications,}
  \city{Beijing}
  \country{China}
}

\author{Xiao Liu}
\email{liuxiao68@bupt.edu.cn}
\affiliation{%
  \institution{Beijing University of Posts and Telecommunications,}
  \city{Beijing}
  \country{China}
}

\author{Guangsheng Yu}
\email{reimusaber@gmail.com}
\affiliation{%
\institution{Independent Researcher}
\city{Sydney}
\country{Australia}}

\author{Bo Du}
 \email{bo.du@griffith.edu.au}
\affiliation{%
\institution{Griffith University,}
 \city{Brisbane}
 \country{Australia}}

\author{Jun Shen}
 \email{jshen@uow.edu.au}
\affiliation{%
  \institution{University of Wollongong,}
  \city{Wollongong}
  \country{Australia}
  }
 

\author{Qiang Wu}
\email{wuqiang@lzu.edu.cn}
\authornote{Qiang Wu is the corresponding author}
\affiliation{%
  \institution{Lanzhou University}
  \city{Lanzhou}
  \country{China}
}

\begin{abstract}
Traffic accidents result in millions of injuries and fatalities globally, with a significant number occurring at intersections each year. 
Traffic Signal Control (TSC) is an effective strategy for enhancing safety at these urban junctures. 
Despite the growing popularity of Reinforcement Learning (RL) methods in optimizing TSC, these methods often prioritize driving efficiency over safety, thus failing to address the critical balance between these two aspects. Additionally, these methods usually need more interpretability.
CounterFactual (CF) learning is a promising approach for various causal analysis fields.
In this study, we introduce a novel framework to improve RL for safety aspects in TSC.
This framework introduces a novel method based on CF learning to address the question: ``What if, when an unsafe event occurs, we backtrack to perform alternative actions, and will this unsafe event still occur in the subsequent period?'' 
To answer this question, we propose a new structure causal model to predict the result after executing different actions, and we propose a new CF module that integrates with additional ``X'' modules to promote safe RL practices.
Our new algorithm, CFLight, which is derived from this framework, effectively tackles challenging safety events and significantly improves safety at intersections through a near-zero collision control strategy.
Through extensive numerical experiments on both real-world and synthetic datasets, we demonstrate that CFLight reduces collisions and improves overall traffic performance compared to conventional RL methods and the recent safe RL model. 
Moreover, our method represents a generalized and safe framework for RL methods, opening possibilities for applications in other domains.
The data and code are available in the github\footnote{https://github.com/AdvancedAI-ComplexSystem/SmartCity/tree/main/CFLight}.
\end{abstract}




\maketitle

\section{Introduction}
\textbf{Motivations.} 
Road accidents have dire consequences, affecting victims, families, and societies worldwide.
The World Health Organization reports an annual loss of 1.4 million lives and 20 to 50 million injuries due to road crashes in 2016~\cite{world2018global}. 
Intersections contribute significantly, causing about one-quarter of traffic fatalities and half of all injuries in the United States every year~\cite{highways_dot_gov}. 
Meanwhile, traffic congestion remains an ever-escalating issue in contemporary urban environments, imposing substantial adverse impacts on various aspects of city life, including economic productivity and time consumption~\cite{pishue2021inrix}. 

In recent years, Deep Reinforcement Learning (DRL)~\cite{drl} has garnered increasing attention by adjusting traffic signal control (TSC) to reduce traffic jams.
By setting appropriate reward functions, such as minimizing vehicle travel time and maximizing traffic flow throughput.
DRL continuously learns and optimizes the control strategies, adapting to changing traffic conditions in real time.
Thus, DRL models make proper decisions regarding traffic light phases based on the current traffic flow state~\cite{mplight,icml}. 
%
%
%
Despite the many advances in RL model for TSC, 
most of the existing DRL models only consider traffic efficiency rather than safety. 
Now, the safe RL methods are proposed to ensure the agent's learning and decisions align with safety constraints to avert hazardous outcomes~\cite{rasheed2020deep}. 
It's vital for real-world RL applications, safe RL holds the potential to tackle unresolved safety challenges in TSC.
CounterFactual (CF) learning for safety is an emerging area of research in the field of AI and aims to improve the safety and robustness of AI systems by considering ``what-if'' scenarios during the training process~\cite{HE20201CF}.
Thus, CF learning provides some valuable tools for studying safety. 
Researchers often conduct CF policy evaluations before deploying any policy to the real world by using CF reasoning to reduce unsafe events. 

The unsafe problems of intersections and the potential of CF learning for safe RL motivate us to develop a safe RL method for TSC to ensure intersection safety and efficiency at the same time.
\begin{figure}[ht]
    \centering
    \includegraphics[width=0.48\textwidth]{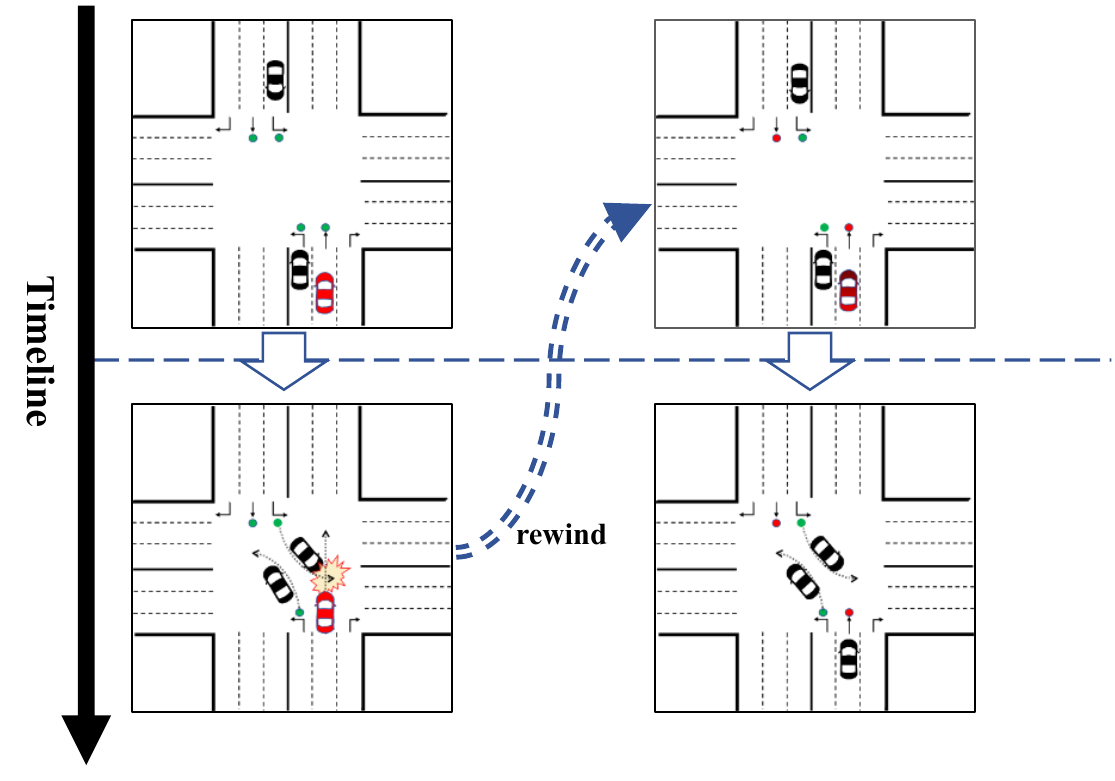}
    \caption{The red vehicle, moving at high speed using permitted-only phase, collides with the opposing left-turning vehicle in permitted-only phase. We rewind to the state of pre-collision, and shift to protected-only phase to reduce collisions by avoiding the approach of hostile vehicles}
    \label{intro}
\end{figure}

\noindent\textbf{Challenges.} 
In the real-world, many intersections use a permitted-only phase, allowing both straight and left turns. While this improves efficiency, it increases the risk of collisions between straight-moving and left-turning vehicles. Alternatively, the protected-only phase, which separates left turns from straight movements, reduces collision risk but significantly decreases traffic efficiency. 
These current safe RL and TSC methods~\cite{gong2020multi,du2023safelight} are still unsatisfactory in reducing the number of collisions and lack interpretability.
Meanwhile, most safe RL methods are developed for specific domains or applications, lacking a universal framework for rapid research and development of specific safe RL solutions. 
In the TSC field, few methods have been proposed to solve safety problems.

\noindent\textbf{Contributions.} In this study, we design a CF framework to improve the safe RL model by answering the CF question: ``What if, when an unsafe event occurs, we backtrack to perform alternative actions, will this unsafe event still occur in the subsequent period?''.
Furthermore, we effectively achieve a balance between traffic efficiency and safety, we have specifically implemented this method in TSC scenarios (as shown in Figure~\ref{intro}) via leveraging this CF framework. Our contributions are summarized as follows:
%
\begin{itemize}
    \item We propose the first instantiation of the causal learning for TSC by answering counterfactual questions to improve the safety of TSC. Specifically, we design structure casual models to predict the result after executing different actions and collect the CF result to enrich the data and improve performance. 
    \item We develop a CF+``X'' safe RL framework that integrates CF components with various ``X'' modules. Based on this framework, we introduce a novel algorithm, CFLight, which effectively addresses safety events and incorporates a new reward function to improve both safety and action interpretability.
    \item Theoretical analysis shows that CFLight, augmented with CF data, converges reliably. Experimental results demonstrate that CFLight significantly reduces collisions, achieving up to a 93.1\% reduction in collision rates compared to 3DQN methods, while maintaining near-zero incident rates.
\end{itemize}
Furthermore, our ``CF+X'' framework serves as a generalized safe RL method that can be applied in various other domains.

\section{Background}

\subsection{Counterfactual learning} 

CF learning, or counterfactual estimation or reasoning, is a fundamental concept utilized across diverse fields such as machine learning, causal inference, and decision-making. 
It has recently gained significant attention leverage unreal data instances~\cite{verma2022counterfactual,lu2020sample,li2021shapley}.
The origin of CF learning traces back to the pioneering work in 1974 ~\cite{rubin1974}, where fundamental concepts such as potential outcomes and causal inference laid the groundwork for subsequent research.
The twin network method ~\cite{pearl2009causality} visually represents reasoning with two networks: one for the factual world and one for the counterfactual (imaginary) world. They share the same structure, but the intervened variables are removed in the counterfactual world. 
Researchers have successfully used this technique in the healthcare sector to estimate the causal effects of medical treatments and interventions ~\cite{Prosperi2020}.
CF learning has also yielded significant breakthroughs in recommender systems, as seen in methods like CF matrix factorization~\cite{xu2020adversarial}.
Furthermore, RL and CF learning amalgamation have impressive performance in optimizing decision-making processes ~\cite{cf-RL-2023,mesnard2020counterfactual,liu2023novel}.

The scope applications of RL expands into safety-critical domains, ensuring the robustness and safety of RL agents~\cite{gu2023review}.
In the early stages of addressing RL safety, primarily directed toward defending against and it reveals such attacks exploit learned policies, leading to unintended and potentially harmful behaviors within RL-based systems~\cite{Sun-2020-AAAI}. 
The event of agent safety within RL has received significant attention, particularly in critical domains like healthcare and autonomous vehicles. A comprehensive framework has been proposed to address the agent safety problem in RL, ensuring agents prioritize safety to mitigate potential risks~\cite{gu2023review}. For TSC, rare applications are focused on safety rather than efficiency.

Structural Causal Models (SCMs) serve as a formal framework in causal inference and statistics to represent and analyze causal relationships among variables.
In a SCM, the causal relationships among variables are explicitly depicted. The set of variables, denoted as $X_1, X_2, ..., X_n$, is associated with structural equations:
\begin{equation} 
\label{SCM}
X_i = f_i(Pa(X_i), U_i),
 \end{equation}
 where:
$Pa(X_i)$ signifies the parents of $X_i$ in the causal graph,
 $f_i(\cdot)$ represents a deterministic function describing how $X_i$ is generated based on its parents and
 $U_i$ stands for the error term, and  it represents the unobserved factors affect $X_i$.
The CF Causal Effect (CCE) is  formulated as the disparity between the potential outcome under an intervention and the observed outcome without the intervention: 
$
\text{CCE} = \tilde{Y} - Y 
$
where $Y$ signifies the observed outcome, and $\tilde{Y}$ represents the CF outcome.
The core objective of CF learning lies in estimating the CF outcome $\tilde{Y}$ or the causal effect, utilizing observed data $\mathcal{D}$ in conjunction with the causal model.

\subsection{Traffic Signal Control}
Due to the continuous advancements in urban transportation, research on TSC has been advancing progressively since its inception.
The fixed-time method~\cite{koonce2008traffic} cannot make optimal signal choices in response to real-time traffic conditions.
Self-Organizing Traffic Lights (SOTL)~\cite{cools2013self} decides phase maintenance or changing by using predefined rules and real-time traffic conditions.
Given constraints of the traditional method, the RL TSC dynamically adjusts signal phases in real time based on current traffic, optimizing performance~\cite{oroojlooy2020attendlight,jin2017intelligent,guo2019reinforcement}. 
CoLight~\cite{CoLight2019} employs a graph attention and significant 
optimization through neighboring intersection data. Advanced-CoLight~\cite{icml} further advances TSC performance. SafeLight~\cite{du2023safelight} integrates a safe component into RL, reducing collision rates.
\begin{figure}[ht]
    \centering    
\includegraphics[width=0.48\textwidth]{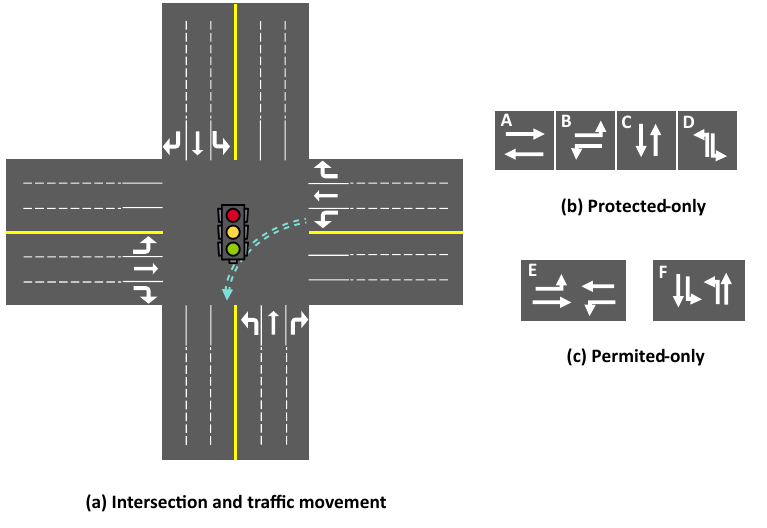}
    \caption{Illustration of a four-way intersection.}
    \label{fig2}
    \vspace{-1em}
\end{figure}
\\
\textbf{Traffic network.} A traffic network is composed of a series of intersections $(I_1,..., I_N)$, with each intersection hosting a variety of roads and each road further accommodating multiple lanes.
\\
\textbf{Traffic movement.} 
Traffic movement is the vehicle motion within an intersection that involves lane changes upon entry and exit (as shown in Figure~\ref{fig2}(a)).
\\
\textbf{Traffic signal phase.} 
A Traffic Signal Phase (TSP) specifies vehicle movements at an intersection during a defined interval. It designates proceeding, yielding, or stopping lanes, each linked to a signal indication. Networks use 2, 4, or $k$ phases, denoted as $a=1,2,...,k$. Figure~\ref{fig2}(b)(c) shows 6 phases. To enhance traffic efficiency, most cities have established a set of regulations that focus on optimizing left-turn scenarios. Several left-turn phasing guidelines were built by Federal Signal Timing Manual~\cite{koonce2008traffic}. These regulations encompass two primary modes:
\begin{itemize}
 \item \textbf{Permitted-only}: allows both straight-ahead and left turns.
 \item \textbf{Protected-only}: permits only straight-ahead movements while prohibiting left turns.
\end{itemize}
Although the ``permitted-only'' traffic signal mode aims to improve traffic efficiency, it inadvertently increases the risk of collisions. This risk arises from faster-moving vehicles on a permissive signal road colliding with vehicles making left turns, posing a critical safety hazard.

To find a solution to this unsafe scenario, the ``protected-only'' signal mode has been designed to restrict left turns on the current road and eliminate conflicts with oncoming traffic. This aims to decrease collisions on risky roads, enhancing overall traffic safety.

\section{Methods}

In this section, we introduce the CF+``X'' framework, which integrates the CF and X modules to enhance the safety of RL.
We describe the creation of CF trajectories and present the safe RL algorithm CFLight for TSC by implementing the CF+``X'' framework. Figure~\ref{fig:framework} shows the overview.

\subsection{CF+``X'' Framework}

The CF+``X'' framework comprises two primary modules: the CF module and the X module.
The CF module employs a SCM model to collect CF trajectories, while the X module integrates a Safe component (e.g., collision penalty in reward) and an Efficient-RL component (e.g., throughput or vehicle waiting time optimization in reward).
\subsubsection{CF Module}
The CF module consists of CF trajectories and the CF objective, which together serve as a vital mechanism for improving RL safety by incorporating CF reasoning into the learning process.
CF trajectories offer additional safety-related experiences from which both the TSC agent and the SCM model can learn.
The CF objective guides the optimization process during training, encouraging safer decision-making in the TSC model.

\textbf{SCM Model.} Assume that the next state $S’$ and the CF result $R$ are governed by the following SCM:
\begin{equation}
\label{SCM1}
S’ = \mathcal{M}_1(S, A, U_s), \quad R = \mathcal{M}_2(S, A, U_r),
\end{equation}
where $S$ and $A$ denote the current state and action, respectively. The variable $U$ represents a noise term, which may include factors such as weather conditions and transmission variability. This noise is an unobserved variable that is assumed to be independent of both $S$ and $A$. The model $\mathcal{M}$ functions as a generative model, capable of predicting unknown outcomes in real-world tasks. The SCM is trained using data stored in the experience buffer of the TSC agent and is used to collect CF trajectories.

\textbf{CF Trajectories.} We outline the process for collecting CF trajectories $(s, a_{cf}, s_{cf}, r_{cf})$.
We begin by recording the states associated with all unsafe events prior to their occurrence, using a simulator denoted as $env$.
Next, the simulation is rewound to the state immediately before each unsafe event. From this point, given alternative CF actions $a_{cf}$, the SCM model $\mathcal{M}$ is used to predict the corresponding CF reward $r_{cf}$ and the resulting CF state $s_{cf}$.
Finally, the outcomes before and after the rewind are compared in order to construct the CF objective.

\begin{figure*}[ht]
\centering
\includegraphics[width=1\textwidth]{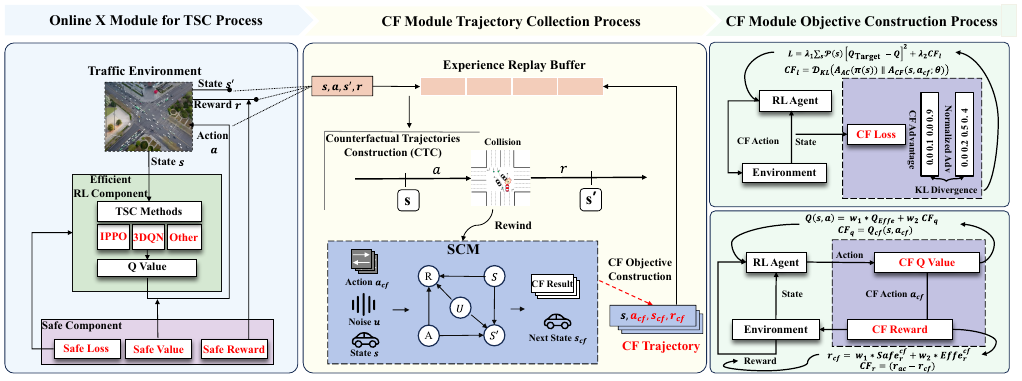}
   
 \caption{The framework of CFLight consists of two modules: the X Module and the CF Module. During the TSC process, pre-collision states $s$ are continuously collected. When a collision occurs, the system rewinds to the recorded pre-collision states and initiates the CF trajectory collection process.
In the CF Trajectories Construction (CTC) step, different actions $a_{cf}$ are evaluated by the SCM model $\mathcal{M}$ to predict potential collisions, the resulting next state $s_{cf}$, and to collect the CF trajectory $(s, a_{cf}, s_{cf}, r_{cf})$.
Finally, the CF Objective is formulated, the CF trajectories are added to the replay buffer, and both the TSC agent and the SCM model are updated.}
 \label{fig:framework}

\end{figure*}
\textbf{CF Objective (Reward, Loss, and $Q$ Value).}
We introduce the concept of a CF objective to enhance safety, defined as:
\begin{equation}
\label{CFReward}
CF = f(R_{CF}, R_{AC}),
\end{equation}
where $R_{CF}$ represents the outcome resulting from the execution of the CF action $a_{cf}$, while $R_{AC}$ denotes the outcome following the execution of the real world intended action.
The function $f$ is designed to measure the distance between $R_{CF}$ and $R_{AC}$.
Additionally, $R_{CF}$ can be incorporated as a separate term in the overall optimization objective to explicitly account for CF considerations during learning.

\subsubsection{X Module}

The X Module consists of two primary components: the Safe component and the Efficient-RL component.
The Safe component is designed to incorporate basic safety considerations into the RL model, while the Efficient-RL component represents the efficiency-driven part of the RL model.
Different configurations of Safe and Efficient-RL components can be flexibly combined to form the integrated X Module.

\textbf{Safe component.} The Safe component focuses on embedding safety considerations into the design of the reward, loss, and $Q$ functions within the RL framework. In the case of Safe Reward ($Safe_{r}$), safety concerns influence the reward function in a manner similar to reward shaping; when the safety model detects an unsafe action, a safety constraint is introduced to guide the learning process. For Safe Loss ($Safe_{l}$), the RL model is enhanced with an additional task by integrating the safe component into the loss function, enabling multi-task learning that optimizes multiple objectives simultaneously. In the case of Safe Value ($Safe_{v}$), the RL model is augmented by incorporating the safe component into the $Q$ function, thereby establishing a multi-task learning framework that concurrently optimizes for both performance and safety.

\textbf{Efficient-RL component.} The Efficient-RL component utilizes advanced RL models aimed at optimizing efficiency. It can either maintain consistency with the original algorithm or adopt novel approaches to improve performance and adaptability.

\subsection{CFLight}
We propose CFLight, a method built upon the “CF+X” framework, designed to enhance both the efficiency and safety of TSC, as illustrated in Figure~\ref{fig:framework}.
\subsubsection{CF Module Implementation}

\textbf{SCM Model.} A Bidirectional Conditional Generative Adversarial Network (BiCoGAN)~\cite{biogan} is used to approximate the SCM model. The core idea is to use a generator network to create fake samples that can deceive the adversarial network. The generator network $\mathcal{M}$ is employed to estimate the next state $S'$ and reward $R$, conditioned on ($S,A,U_s,U_r)$. Encoder is an inference mapping from $S'$ and $R$ to ($S, A, \hat{U_s}, \hat{U_r}$). 
 \begin{equation} 
\label{Ecoder} 
\begin{aligned} 
S, A, \hat{U_s}, \hat{U_r}& = E(S',R),\quad 
\hat{S'},\hat{R}=\mathcal{M}(S,A,U_s,U_r)
\end{aligned}
 \end{equation}
Using the data from the experience buffer, the SCM model is updated using the following formula:
\begin{equation} 
\label{SCM_IMP} 
\begin{aligned} 
\mathcal{L} & =\min _{\mathcal{M},E} \max _D V(D, \mathcal{M}, E)=\mathcal{L}_{DE}+\mathcal{L}_{DM}+\beta \mathcal{L}_{\mathcal{SR}}+\lambda \mathcal{L}_{mono},
\end{aligned}
 \end{equation}
where
$\mathcal{L}_{DE} = \mathbb{E}[\log D(S, A, R, S', \hat{U}_s, \hat{U}_r)]$ and
$\mathcal{L}_{DM} = \mathbb{E}[\log D(S, A, \\ \hat{R}, \hat{S}', U_s, U_r)]$
are GAN-based loss functions used to optimize the discriminator, encoder, and generator models, helping to better distinguish between real and generated transitions.
The loss $\mathcal{L}_{SR} = | (S', R) - (\hat{S}', \hat{R}) |_2$ encourages the generator to produce states and rewards that more closely approximate the ground truth, using the $L_2$ norm distance.
The monotonicity penalty,
$\mathcal{L}_{\text{mono}} = \sum_l \sum_{i,j} \max(0,\\ -W_{l,i,j})$,
enforces the monotonicity of the generator $\mathcal{M}$, where $l$ denotes the layer index and $W_{l,i,j}$ represents the weights of $\mathcal{M}$.
The hyperparameters $\lambda$ and $\beta$ control the trade-off among the loss components.

\begin{theorem}
\label{theorem_1}
 Suppose $R$ and $S'$ satisfy the below causal model:
 \begin{equation} 
\label{theorom}
S' = \mathcal{M}_1(S,A,U_s),\text{   } R=\mathcal{M}_2(S,A,U_r),
 \end{equation}
 where $U_s,U_r\Perp (S,A)$, $\mathcal{M}_1,\mathcal{M}_2$ is smooth and monotonic in $U_s,U_r$ for fixed value of $S,A$.  Suppose we have observed $(S,A,R,S')$, then for the CF action $a_{cf}$, the following CF outcome: 
  \begin{equation} 
\label{theorom1}
S'_{A=a_{cf}},R_{A=a_{cf}} \mid S=s, A=a, S'=s', R=r,
 \end{equation} is identifiable.
\end{theorem}
\textbf{Proof:} Given $S' = \mathcal{M}_1(S, A, U_s)$ and the observed values $S = s$, $A = a$, and $S' = s'$, and assuming that $\mathcal{M}_1$ is  monotonic in $U_s$, we can invert the function to obtain $\hat{u}_s = \mathcal{M}_1^{-1}(s', s, a)$. Setting $A = a_{cf}$, and noting that $U_s \Perp (S, A)$, the noise term $\hat{u}_s$ remains unchanged. According to Theorem 1 in~\cite{lu2020sample}, we can then compute the CF next state as $S'_{A = a_{cf}} = \mathcal{M}_1(s, a_{cf}, \hat{u}_s)$. By the same reasoning, the CF reward is given by $R'_{A = a_{cf}} = \mathcal{M}_2(s, a_{cf}, \hat{u}_r)$. Therefore, the CF outcomes are identifiable.

According to Theorem~\ref{theorem_1}, we obtain the CF outcomes $s_{cf} = \mathcal{M}_1(s, a_{cf}, r_s)$ and $r_{cf} = \mathcal{M}_2(s, a_{cf}, r_r)$. This indicates that by changing to the CF action $a_{cf}$, we can generate the CF next state $s_{cf}$ and the CF reward $r_{cf}$. Finally, we use the loss function defined in Equation~\ref{SCM_IMP} to optimize the SCM model $\mathcal{M}$.

\textbf{CF Trajectories.} 
Algorithm~\ref{algo:ctc} shows the CF trajectories construction (CTC) step.
CF thinking infers what would have happened if an agent had taken a different action. Given a sample $(s, a_{cf}, u)$, a CF action $a_{cf}$ is randomly drawn from the action space and, together with state $s$ and noise $u$, passed to the generator $\mathcal{M}$ to predict the CF reward $r_{cf}$ and next state $s_{cf}$.

\textbf{CF Objective.} We design different instances of the function $f$ in Equation~\ref{CFReward} for reward, loss, and $Q$ value, respectively, to define the CF Objective.
Here, $r_{ac}$ and $r_{cf}$ represent the rewards obtained by executing the action and the CF action, respectively. These different CF objectives encourage the agent to learn safer strategies.
To emphasize the difference between real-world and CF collisions, we define the following reward-based CF objective:
\begin{equation}
\label{CF-reward}
CF_{r} = r_{cf} - r_{ac},
\end{equation}
where $r_{ac} = w_1 \cdot Safe_r + w_2 \cdot Effe_r$ captures the reward composed of the number of collisions ($Safe_r$) and the total vehicle waiting time ($Effe_r$), both observed after executing the pre-collision action $a$.
In the computation of the CF reward, $r_{cf} = w_1 \cdot Safe^{cf}_r + w_2 \cdot Effe^{cf}_r$, various CF actions $a_{cf}$ are explored to determine the number of CF collisions $Safe^{cf}_r$ and the efficiency measure $Effe^{cf}_r$.

\begin{algorithm}[t]
\caption{Counterfactual Trajectories Construction (CTC)}
\label{algo:ctc}
\SetAlgoLined
\SetKwFunction{CTC}{CTC}
\SetKwProg{Fn}{Function}{:}{}
\Fn{\CTC{SCM models $\mathcal{M}_1$ and $\mathcal{M}_2$, Pre-collision state $s$, Real-world reward $r_{ac}$}}{
  $trajectory = [\quad ]$\;
  \For{$a_{cf} = 1$ \KwTo $n$}{
    Rewind pre-collision state $s$ and initialize noise $u_s,u_r$\;
    Predict $s_{cf}=\mathcal{M}_1(s,a_{cf},u_s)$, $r_{cf}=\mathcal{M}_2(s,a_{cf},u_r)$\;
    Construct $CF_r$ by Equation~\ref{CF-reward}, $CF_l$ by Equation~\ref{CF-action} and $CF_{q}$ by Equation~\ref{CF-collision}\;
    $trajectory.append(s,a_{cf},s_{cf},r_{cf})$\;
  }
  \Return{$trajectory$}\;
}
\end{algorithm}

This reward quantifies the causal effect of executing the CF action $a_{cf}$. Assuming that efficiency remains constant before and after the intervention, and that collisions decrease from 2 (in the real-world observation) to 0 (in the CF scenario), the resulting reward difference becomes $CF_r = 0 - (-2) = 2$.
Such a CF trajectory $(s, a_{cf}, r_{cf})$ with an improved reward of 2 can substantially enhance the model’s learning performance for safety improvement, and conversely, a negative reward could indicate actions to avoid.
\begin{algorithm}[!ht]
\caption{CFLight Training (CFLight-R, CFLight-Loss, CFLight-Q)} 
\label{algo:CFLight}
\SetAlgoLined
\KwIn{Input epoch times $K$, replay buffer $D,D_{CF}$, max steps $maxSteps$, pre train steps $preTrainSteps$, random epsilon $eps$, simulator $env$ and SCM models $\mathcal{M}$.}
\For{$i = 1$ \KwTo $K$}{
    $s = env.getState()$\;
    Initialize efficient RL and safe component and get $Q$ network\;
    $totalSteps = 0$\;
    \For{$j = 1$ \KwTo $maxSteps$}{
        $totalSteps += 1$\;
        \eIf{random number $< eps$}{
            Explore random action\;
        }{
            Choose action $a$ based on $Q$ network\;
        }
        In real world, execute the action $a$ by $env$, get $Safe_{r}$, real world trajectory $(s, a, r, s')$ and add it into replay buffer $D$\;
        
        \If{$totalSteps > preTrainSteps$}{
            Sample trajectories from $D$ and $D_{CF}$\;
            Update $Q$ by Equation~\ref{equation:bellman}(CFLight-R), Equation~\ref{safelightLoss}(CFLight-Loss) or Equation~\ref{synQ} (CFLight-Q)
        }
    }
    \For{i \% trainGap == 0}{
    $D_{CF}.clear()$\;
    Update $\mathcal{M}$ by Equation~\ref{SCM_IMP}\;

    \If{collision}{
            Go to CF world to collect CF trajectories $(s, a_{cf}, r_{cf}, s_{cf})$ according to CTC function\;
            Add CF trajectories into replay buffer $D_{CF}$\;
        }
    }
}
\end{algorithm}

The distributional distance between the CF action and the  action is formalized by:
\begin{equation}
\label{CF-action}
CF_{l} = D_{KL}(A_{AC}(\pi_H(s)) || A_{CF}(s, a_{cf}; \theta)),
\end{equation}
where a collision occurs upon executing a real-world action, and we obtain the unsafe advantage distribution $A_{AC}$ generated by the policy $\pi_H(s)$, which takes the current traffic state as input and determines whether an action is unsafe.
We capture collision-related information, trace it back to the CF scenario, and explore alternative safe actions to derive the safe advantage distribution $A_{CF} = Q(s, a_{cf}; \theta) - V(s)$, measuring the discrepancy via cross-entropy.

The occurrence of a collision in the CF world also introduces a novel objective aimed at risk mitigation. We explore the CF action $a_{cf}$ and evaluate its corresponding value $CF_q$ as:
\begin{equation}
\label{CF-collision}
CF_{q} = Q_{cf}(s, a_{cf}),
\end{equation}
where $Q_{cf}$ is a network used to assess the safety value of the CF action $a_{cf}$. This value contributes to the overall Q-value computation.
The CF reward $CF_{r}$ is used to update $Q_{cf}$ based on the Bellman Equation~\ref{equation:bellman}.

\begin{table*}[ht]
\centering
\caption{Performance of all methods on synthetic and real-world datasets. Green colors show the percentage improvement when adding the CF module. Red colors indicate the percentage of degradation.}
\label{table:overall}
\resizebox{\linewidth}{!}{
\begin{tabular}{|c|c|c|c|c|c|c|c|c|c|c|} 
\hline
\multicolumn{11}{|c|}{Synthetic — Acyclic Action}                                                                                                                                                                                                                                                                                                                                                                                                                                             \\ 
\hline
                  & SafeLight-Act                                                                                                        & CPO                                                                        & Syn-R            & CFLight-R (Improved)                                   & SafeLight-Loss   & CFLight-Loss (Improved)                       & Syn-Q           & CFLight-Q (Improved)                             & 3DQN                            & Fixed-time      \\ 
\hline
Average Delay (s) & \begin{tabular}[c]{@{}c@{}}\textcolor[rgb]{0.2,0.2,0.2}{4.73}\\\textcolor[rgb]{0.2,0.2,0.2}{$\pm$1.04}\end{tabular}  & \textcolor[rgb]{0.2,0.2,0.2}{22.16}\textcolor[rgb]{0.2,0.2,0.2}{$\pm$3.23} & 9.25$\pm$1.03    & 8.67$\pm$1.07~\color{green!70!black}(+6.3\%)                   & 30.29$\pm$1.28   & 29.89$\pm$2.15\color{green!70!black}(+1.3\%)          & 8.54$\pm$0.14   & 8.87$\pm$1.11\color{red}(-3.8\%)                         & \textbf{4.24}$\pm$\textbf{0.13} & 31.22$\pm$1.34  \\ 
\hline
Throughput        & \begin{tabular}[c]{@{}c@{}}1775.3\\$\pm$52.35\end{tabular}                                                           & 1760.2$\pm$43.31                                                           & 1784.8$\pm$50.03 & \textbf{1802.8$\pm$80.25}~\color{green!70!black}(+1.0\%)       & 1106.2$\pm$30.05 & 1107.2$\pm$30.05\color{green!70!black}(+0.01\%)       & 1753$\pm$40.09  & 1781.3$\pm$40.07\color{green!70!black}(+1.59\%)\textbf{} & 1785.6$\pm$0.04                 & 1782$\pm$0.05   \\ 
\hline
Collision Count   & \begin{tabular}[c]{@{}c@{}}\textcolor[rgb]{0.2,0.2,0.2}{15.9}\\\textcolor[rgb]{0.2,0.2,0.2}{$\pm$6.21}\end{tabular}  & \textcolor[rgb]{0.2,0.2,0.2}{13.2}\textcolor[rgb]{0.2,0.2,0.2}{$\pm$3.53}  & 3.5$\pm$1.16     & 2.5$\pm$1.05\color{green!70!black}(+28.6\%)                    & 1.8$\pm$0.15     & \textbf{1.3$\pm$0.16 \color{green!70!black}(+27.7\%)} & 5.7$\pm$4.33    & 4.6$\pm$2.12 \color{green!70!black}(+19.2\%)             & 19$\pm$0.15                     & 7.6$\pm$0.16    \\ 
\hline
\multicolumn{11}{|c|}{Real-world — Cologne, Acyclic Action}                                                                                                                                                                                                                                                                                                                                                                                                                                   \\ 
\hline
Average Delay (s) & \begin{tabular}[c]{@{}c@{}}\textcolor[rgb]{0.2,0.2,0.2}{23.32}\\\textcolor[rgb]{0.2,0.2,0.2}{$\pm$3.12}\end{tabular} & 44.06$\pm$8.24                                                             & 12.59$\pm$3.14   & 20.27$\pm$3.13 \color{red}(-61.1\%)                            & 66.86$\pm$1.19   & 47.74$\pm$3.56 \color{green!70!black}(+40.0\%)        & 16.20$\pm$5.18  & 22.28$\pm$6.16\color{red}(-37.5\%)                       & \textbf{5.54}$\pm$\textbf{1.12} & 67.25$\pm$3.25  \\ 
\hline
Throughput        & \begin{tabular}[c]{@{}c@{}}2013\\$\pm$0.04\end{tabular}                                                              & 2014.6$\pm$0.04                                                            & 2014$\pm$0.07    & \textbf{2014$\pm$0.05}                                 & 1733.5$\pm$5.03  & 2014.7$\pm$0.04 \color{green!70!black}(+16.2\%)       & 2014.4$\pm$0.05 & 2014.8$\pm$0.04\color{green!70!black}(+0.01\%)           & 2014$\pm$0.05                   & 2015$\pm$0.04   \\ 
\hline
Collision Count   & \begin{tabular}[c]{@{}c@{}}\textcolor[rgb]{0.2,0.2,0.2}{13.2}\\\textcolor[rgb]{0.2,0.2,0.2}{$\pm$6.32}\end{tabular}  & 41.3$\pm$8.56                                                              & 20.3$\pm$6.21    & \textbf{5.3}$\pm$\textbf{4.56} \color{green!70!black}(+73.8\%) & 13.4$\pm$5.34    & 13.3$\pm$4.23 \color{green!70!black}(+0.01\%)         & 13.06$\pm$2.51  & 9.9$\pm$3.34\color{green!70!black}(+24.1\%)\textbf{}     & 40.07$\pm$4.13                  & 35.1$\pm$3.32   \\
\hline
\end{tabular}
}
\end{table*}

\subsubsection{CFLight}

We introduce CFLight-R, a variant of the CFLight algorithm based on the “CF+X” framework. The CF Objective utilizes $CF_r$, the Safe module incorporates $Safe_r$, and the Efficient-RL module uses $Effe_r$, which represents the total vehicle waiting time.
The components of CFLight-R can be succinctly represented as a combination:
$CF_r + Safe_r + Effe_r$.

\begin{itemize}
    \item \textbf{State.} The state $S$ contains the vehicle speed and location. 
    \item \textbf{Action.} The action $A$ is a combination of the \textbf{permitted-only} and \textbf{protected-only} phases.
    \item \textbf{Reward.} The reward $r$ takes the value $R_{AC}$ in the real world and $CF_r$ in the CF scenario.
\end{itemize}


Finally, we update the network through 3DQN and $Q$ is updated according to the Bellman function:
\begin{equation}
\label{equation:bellman}
Q(s,a) \longleftarrow Q(s,a) + \alpha \left [ R + \gamma\max_{a'\in {a}}Q'(s',a') -Q(s,a) \right ]
\end{equation}

Using CF data augmentation, the following lemma ensures that the Q-function will achieve the optimal value~\cite{lu2020sample}:
\begin{lemma}
 Given the transition dynamics, $Q$-learning on the counterfactually augmented data set converges with probability one to the optimal value function $Q^*$, as long as the state and action spaces are finite, and the learning rate $\alpha_t$ satisfies $\sum_t \alpha_t=\infty$ and $\sum_t \alpha_t^2<\infty$.
\end{lemma}

Algorithm~\ref{algo:CFLight} shows a specific implementation based on CF module $CF_{r}$,
X module ($Safe_{r}$ as safe component; $Effe_r$ as efficient-RL component).
The specific CFLight easily accommodates different CF and X modules.
These methods are detailed in \textbf{Appendix ~\ref{detail_algo}}.

\section{Experiments}
In this section, we concentrate on non-periodic acyclic phase switching and extensively experiment with synthetic and real-world datasets.
These experiments are conducted in the microscopic traffic
simulator, SUMO~\cite{SUMO}. We set the collision rate to 40\% under time-variant traffic flow. Detailed datasets, network structure and hyperparameters are in \textbf{Appendix Table~\ref{table: hyperparameter}}.

\begin{figure*}[ht]
    \centering
    \begin{minipage}{0.32\textwidth}
        \centering
        \includegraphics[width=\textwidth]{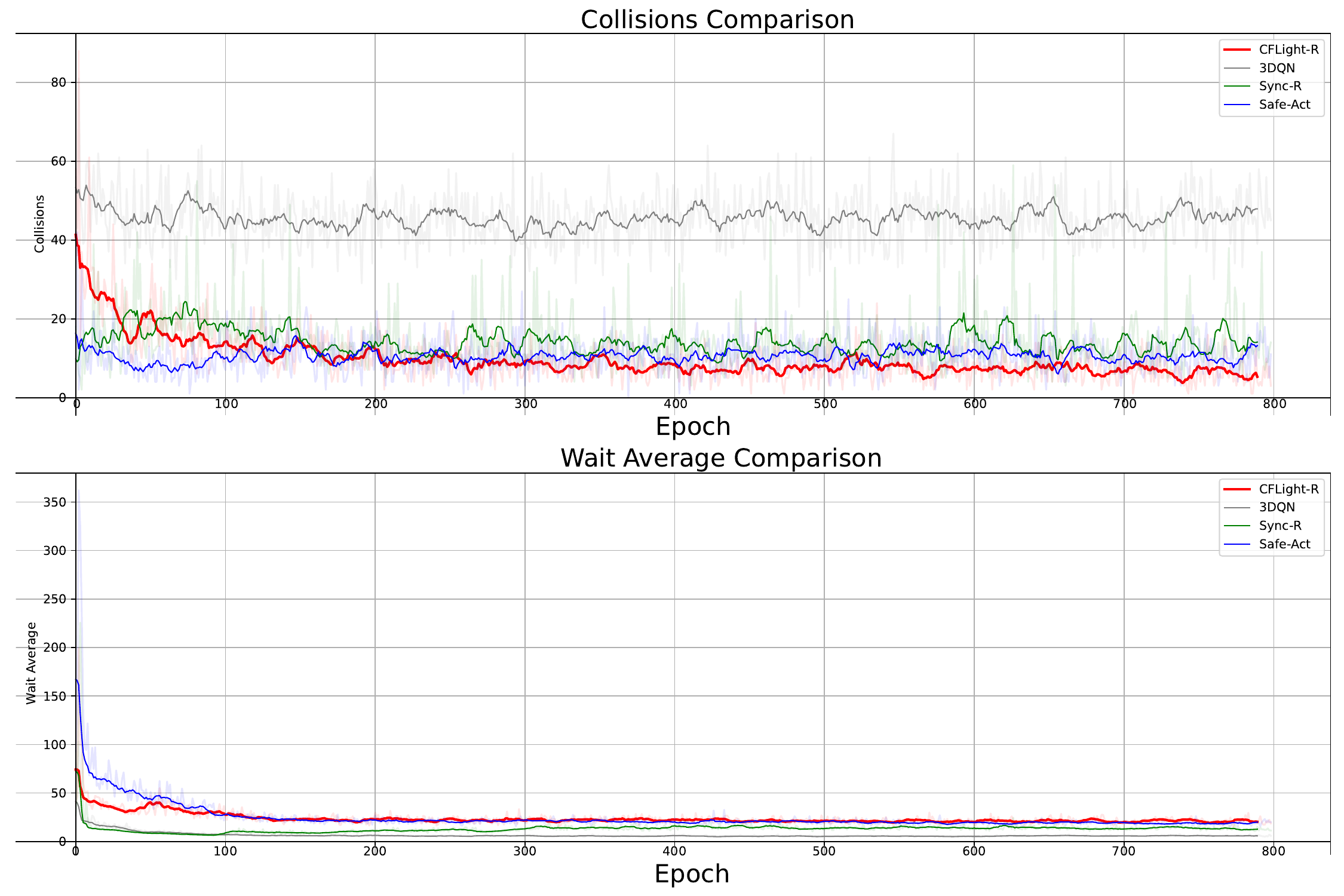}
        \caption*{CFLight-R}
        \label{fig:compare_r}
    \end{minipage}
    \hfill
    \begin{minipage}{0.32\textwidth}
        \centering
        \includegraphics[width=\textwidth]{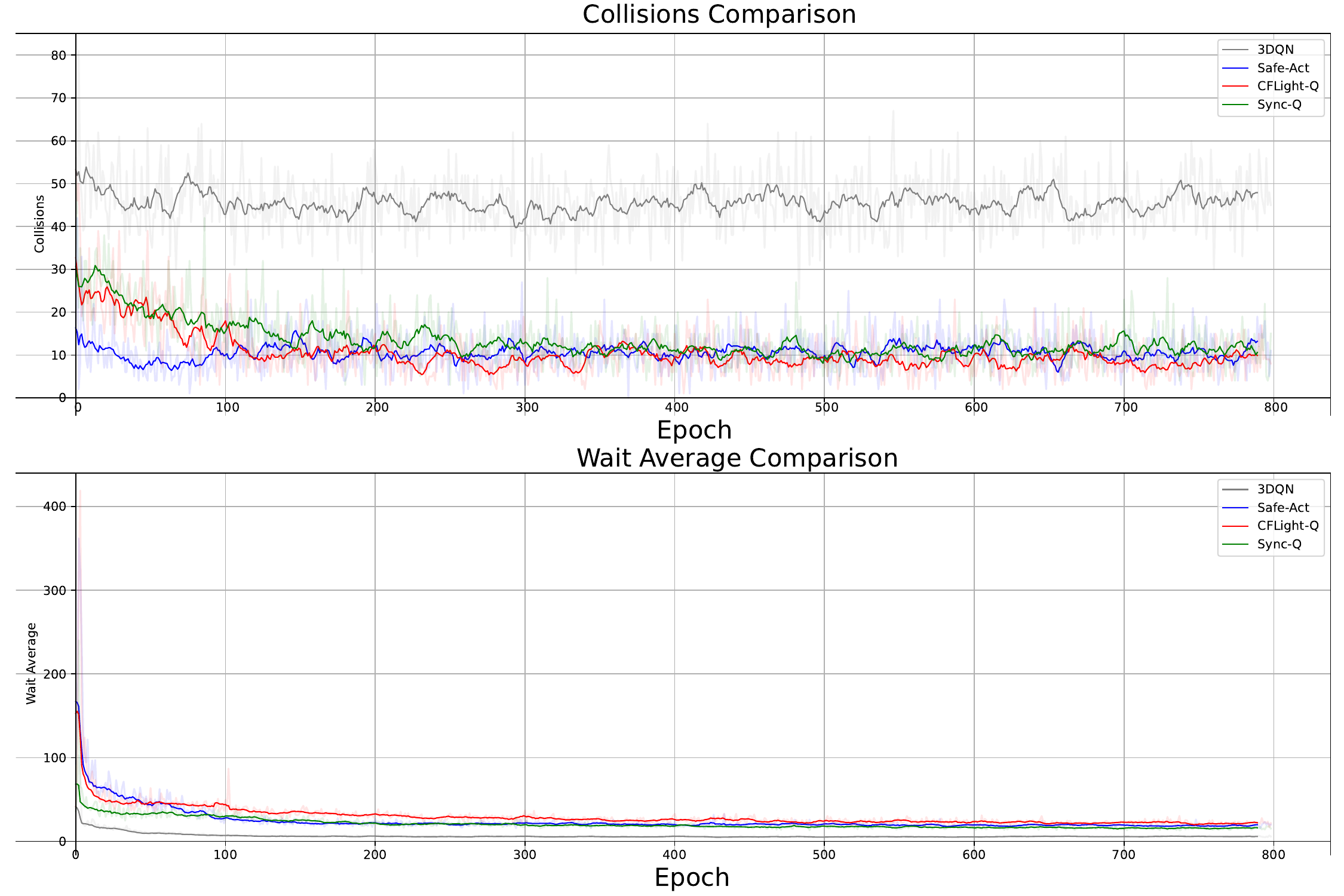}
        \caption*{CFLight-Q}
        \label{fig:compare_q}
    \end{minipage}
    \hfill
    \begin{minipage}{0.32\textwidth}
        \centering
        \includegraphics[width=\textwidth]{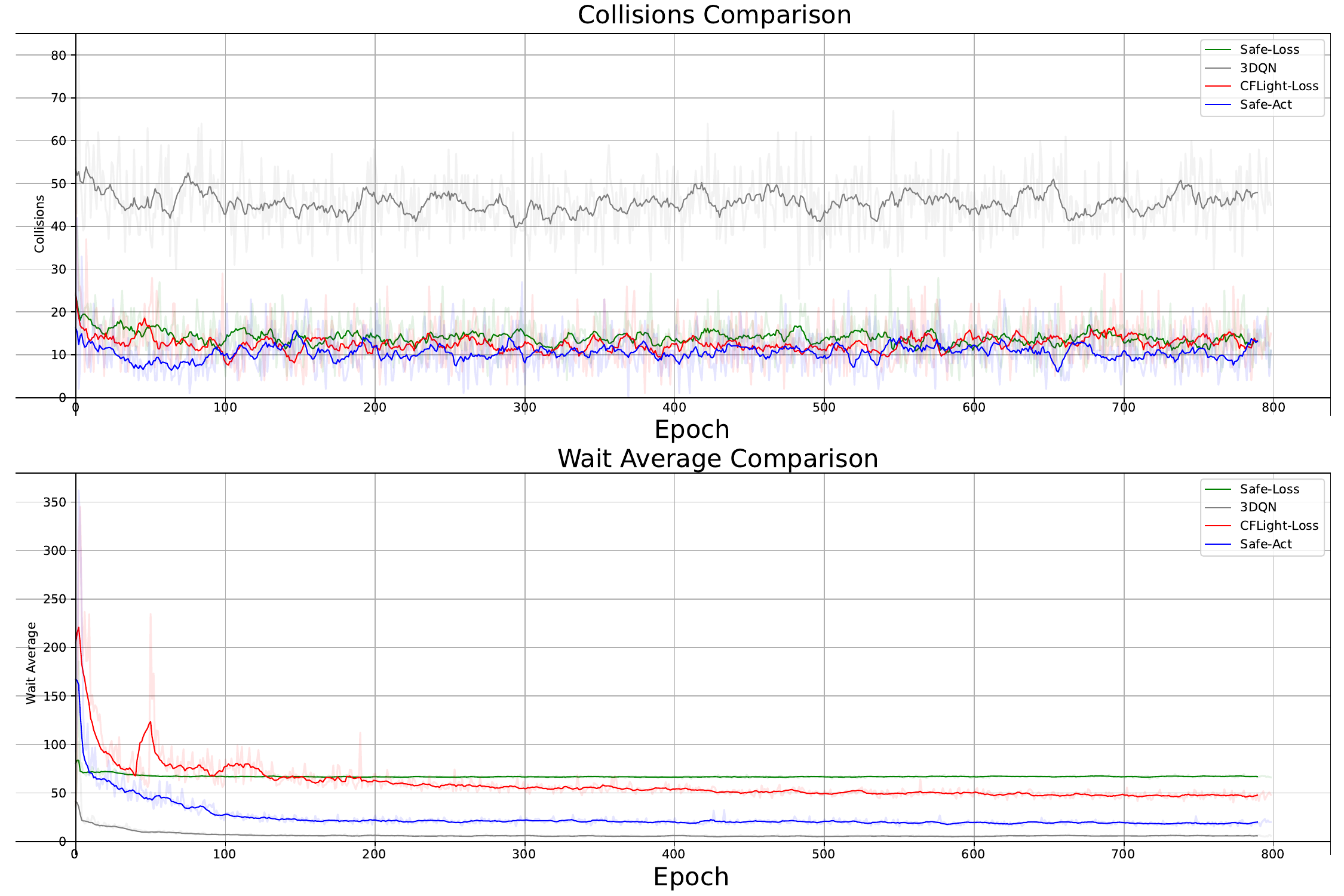}
        \caption*{CFLight-Loss}
        \label{fig:compare_loss}
    \end{minipage}
    \caption{The figures show the traning curve for safety and efficiency for using CFLight in the Cologne dataset.}
    \label{training}
\end{figure*}
\subsection{Datasets}
We experiment with a synthetic intersection~\cite{liang2019deep} and a real-world intersection~\cite{mei2022libsignal}.

\textbf{Synthetic:} The synthetic intersection comprises four methods and twelve one-way vehicular movements with synthetic traffic flow adopted into SUMO. 

\textbf{Real-world:} The real-world dataset is derived from the Cologne, Germany intersection, incorporating real traffic patterns into the SUMO environment. This intersection features 8 approaching lanes. In both cases, the SUMO simulation environment assumes accurate vehicle detection, consistent deceleration rates (modifiable as needed), and permitted left turns within the timing plan.

\subsection{Compared Methods}
In this study, we compared four types of RL methods for intersection control. These methods include:

\textbf{Traditional method:} Fixed-time~\cite{koonce2008traffic}. 

\textbf{RL methods:} 3DQN~\cite{liang2019deep}, IPPO, Syn-R~\cite{khamis2014adaptive}, Syn-Q~\cite{gong2020multi}, CPO~\cite{achiam2017constrained} and SafeLight-Loss~\cite{du2023safelight}.

\textbf{Our methods:} CFLight-1 corresponds to $CF_{r}+Safe_{r}+RL_{3DQN}$. CFLight-2 corresponds to $CF_{l}+Safe_{l}+RL_{3DQN}$. CFLight-3 corresponds to $CF_{q}+Safe_{v}+RL_{3DQN}$.  

All methods share the same action space, which includes both protected-only and permitted-only phases.
More extended methods showed in \textbf{Appendix A}. We evaluate these TSC methods from the perspectives of efficiency and safety. 
The efficiency measure includes Average Delay (average vehicle waiting time) and the number of vehicles throughput within a period of time, while the safety measure is the sum of collisions over time.

\subsection{Results} 

We average the results from the last 10 rounds across two benchmark datasets, with key outcomes presented in Table~\ref{table:overall}. On the synthetic dataset, CFLight-Loss reduces collision rates by 93.1\% compared to 3DQN, outperforming current state-of-the-art (SOTA) methods. On the real-world dataset, the three CFLight variants achieve an average 32.6\% reduction in collisions over non-CF methods, with CFLight-R emerging as the top-performing solution based on collision metrics.

Figure~\ref{training} illustrates the training curves of the model after incorporating the CF module. It can be observed that the training curve for collisions with the CF module shows an improvement compared to the one without the CF module, demonstrating the effectiveness of our approach.


\subsection{Safety and Efficiency Trade-off}
In this section, we illustrate the influence of adjusting weights on both safety and efficiency in synthetic dataset. Leveraging the CFLight-1 algorithm, we conduct 800 training rounds and establish the benchmark value by averaging the results from the last 10 rounds, as detailed in Figure~\ref{fig2}. The outcome reveals the emergence of a Pareto optimal surface. Importantly, a harmonious equilibrium is attained when the weights assigned to safety and efficiency closely align. These insights are derived from testing on a synthetic dataset using CFLight-R.

\begin{table}[t]
\centering
\caption{Ablation experiment in Real-world datasets}
\label{table:abalation}
\scalebox{0.6}{
\begin{tabular}{|c|c|c|c|c|c|c|} 
\hline
                 & \multicolumn{6}{c|}{Real-world Cologne Acyclic Action}                                    \\ 
\hline
                 & $CF_r$      & $CF_r$+$RL_3DQN$ & 3DQN       & $Safe_r$    & $CF_r$+$Safe_r$ & Fixed-time  \\ 
\hline
Average Delay(s) & 412.02±4.24 & 10.41±3.96       & 5.54±1.12  & 706.57±5.21 & 710.70±4.16     & 67.25±3.25  \\ 
\hline
Throughput       & 1684.2±0.13 & 2014±0.04        & 2014±0.05  & 1072.1±0.06 & 1472.9±0.05     & 2015±0.04   \\ 
\hline
Collisions       & 21.8±4.16   & 33.5±5.14        & 40.07±4.13 & 6.9±4.17    & 4.4±3.15        & 35.1±3.32   \\
\hline
\end{tabular}}
\vspace{-1em}
\end{table}

\subsection{Ablation Analysis}
\label{abalation analysis}
We conduct ablation experiments on real datasets involving $CF_{r}$, $CF_{r}$+3DQN, 3DQN, $Safe_r$ and $CF_{r}$+$Safe_r$ configurations to assess the efficacy of the CF module. 
Notably, employing only the CF reward $CF_{r}$ leads to a 45.5\% reduction in collision incidents compared to 3DQN, but with a huge increase in waiting time because it works without optimization in waiting time.
By using $CF_{r}$+$RL_{3DQN}$ method, collision rates decrease by 16.3\%, and waiting times improve significantly compared to the standalone CF module $CF_{r}$.  
Compared to the $Safe_r$ method, CF+$Safe_r$ module the collisions reduce about 36.2\% which shows the validity of our CF module (as shown in Table~\ref{table:abalation}).

\subsection{Case Study}
In this section, we present a visualization of the next state and reward predicted by the SCM $\mathcal{M}$ before and after a collision event. The traffic signal phase prior to the collision is set to permitted-only, and we evaluate the state and reward variations when the SCM predicts outcomes for a protected-only phase. The state is modeled as a $60 \times 60$ matrix, with the $x$- and $y$-axes representing vehicle coordinates and the heatmap values indicating normalized vehicle speeds. The CF reward is computed as $CF_r = r_{cf} - r_{ac}$. In the third subplot on the right side of the figure, the upper half of the reward plot reflects safety (higher values improve safety) and the lower half represents efficiency. The results reveal that adopting the protected-only phase, as predicted by the SCM, significantly increases the safety reward, highlighting its effectiveness in improving the learning of the CF trajectory. Moreover, the CF next states closely resemble the real-world next states in structure, validating the robustness and accuracy of our proposed method.

\begin{figure}[t]
\centering
\includegraphics[width=0.48\textwidth]{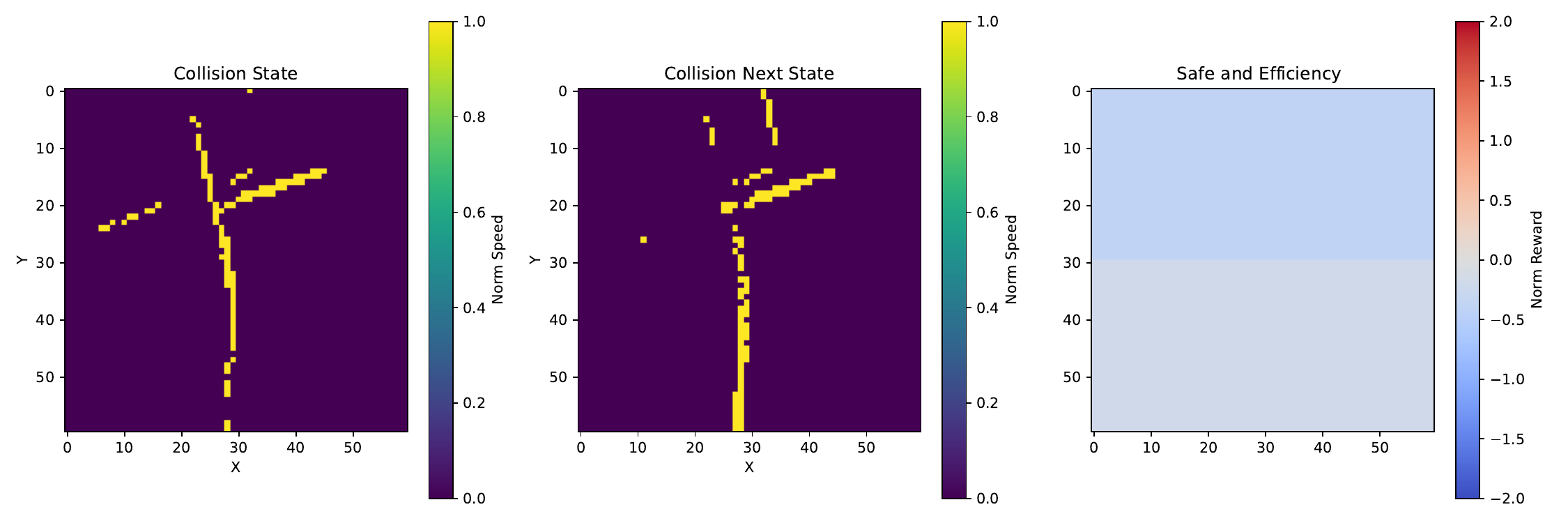}
\caption{In the real world collision case, execute the permitted-only ($a=0$) phase for state, next state, and reward. The third subgraph shows safety (top) and efficiency (bottom).}
\label{fig:cf}
\end{figure}
\begin{figure}[t]
\centering
\includegraphics[width=0.48\textwidth]{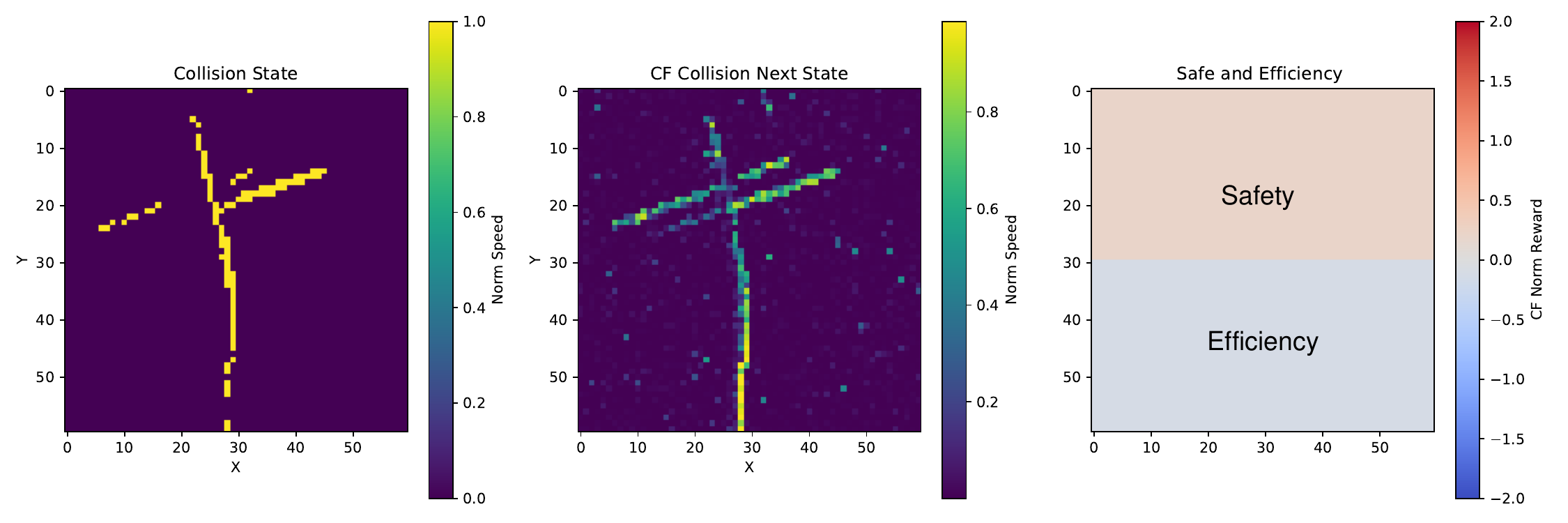}
\caption{In the CF world collision case, execute the protected-only ($a_{cf}=3$) phase for state state, next state and reward $CF_r$ predict by SCM $\mathcal{M}$. The third subgraph shows safety (top, the higher the better) and efficiency (bottom)}
\label{fig:cf}
\end{figure}
\section{Discussion}

\begin{figure}[ht]
    \centering
\includegraphics[width=0.48\textwidth]{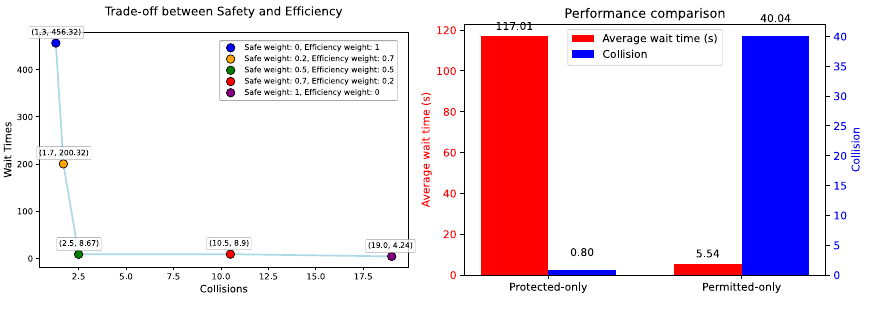}
    \caption{The left image shows the safety-efficiency trade-off, while the right compares protected-only and permitted-only phases based on average waiting time and collision counts in the Cologne dataset.}
    \label{total}
\end{figure}
\textbf {(1) How does the CF module's SCM with GAN approximation operate and what are its benefits?} In CFLight, to realistically simulate the aftermath of traffic collisions, we configure the SUMO simulator to introduce a delay after a collision vehicles involved are briefly held before being rerouted to their destinations. This setup leads to temporary congestion and increased waiting time, aligning with real-world traffic scenarios. Moreover, CFLight introduces a SCM and employs a BiCoGAN to generate counterfactual trajectories under alternative actions, enabling counterfactual data augmentation. This technique creates diverse and realistic hypothetical scenarios, enhancing data diversity and model generalization. As a result, it improves sample efficiency and overall performance without requiring additional real-world data, and comes with theoretical convergence guarantees (Lemma 2).

\noindent\textbf{(2) Why not just use protected-only phases?} We do experiments using only protected-only in the real-world Cologne (1 intersection) dataset as shown in Figure~\ref{total}, and we see that passage efficiency is greatly reduced. 
The number of collisions is greatly increased by using only permitted-only phase. 
In the real world, we observe that there are only two lanes on each road and each lane contains two directions, and the protected-only phase only allows passage in one direction, which may be the reason for the reduced efficiency.
The original intention of our CFLight design is to flexibly switch between these two phases, ensuring both efficiency and safety in traffic.\\

\begin{figure}[h]
\centering
\includegraphics[width=0.45\textwidth]{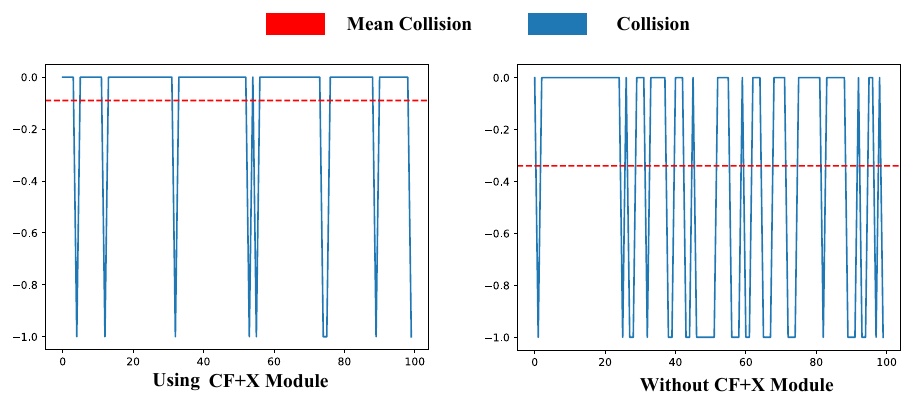}
\caption{The collision result of using our proposed CF+X and without using CF+X under the lane change scenario. Our proposed method improved 73.5\% performance in terms of average collisions on 100 rounds}
\label{fig: lanchange}
\end{figure}

\noindent\textbf{(3) Are there any other scenario experiments for our proposed method?} We conduct lane-changing experiments in an autonomous driving setting to evaluate the CF+X framework. In this task, a vehicle must change across two lanes while assessing collision risk. The state includes the speed of nearby vehicles, the action indicates whether the vehicle changes lanes, and the reward reflects whether a collision occurs. We train an RL agent using CF+X with $CF_r$ for 100 rounds and compare it to a baseline without CF+X. As shown in Figure~\ref{fig: lanchange}, the left graph shows that CF+X reduces the average collision rate to 0.09, compared to 0.34 without the framework, representing a 73.5\% reduction in collisions.

\noindent\textbf{(4) Can our method be applied to multiple intersections?} We used the TSC benchmark~\cite{ault2021reinforcement} datasets with multiple intersections (3, 8 and 100 intersections) and conduct experiments with IPPO as the backbone, CFLight uses $CF_r$ as the CF objective. Our results, as shown in Table~\ref{table:100sync} and Table~\ref{table:ippo}, indicate that CFLight effectively enhances performance, demonstrating the effectiveness of our method.
\begin{table}
\centering
\caption{CFLight performance on 100 intersections of synthetic dataset.}
\label{table:100sync}
\scalebox{1}{
\begin{tabular}{|c|c|c|c|} 
\hline
\multicolumn{4}{|c|}{Synthetic 100 Intersections Acyclic Action}    \\ 
\hline
                  & CFLight-R~       & Syn-R    & Fixed-time  \\ 
\hline
Average Delay (s) & \textbf{3098.90} & 3162.37 & 32708.37    \\ 
\hline
Throughput        & 6831    & \textbf{6852}    & 7065        \\ 
\hline
Collision Count   & \textbf{47}      & 55      & 699         \\
\hline
\end{tabular}
}
\end{table}

\begin{table}
\renewcommand\arraystretch{1}
\centering
\caption{Multiple intersection experiments in Cologne3 and Cologne8.}
\label{table:ippo}
\scalebox{0.9}{
\begin{tabular}{|c|c|c|c|} 
\hline
                 & \multicolumn{3}{c|}{Real-world Cologne3 Acyclic Action}  \\ 
\hline
                 & CFLight-R & IPPO         & Fixed-time   \\ 
\hline
Average Delay(s) & 15.21±2.16                 & 7.37±3.13    & 65.9±5.32    \\ 
\hline
Collisions       & 14.07±3.14                 & 30.07±3.65   & 22±4.13      \\ 
\hline
                 & \multicolumn{3}{c|}{Real-world Cologne8 Acyclic Action}  \\ 
\hline
Average Delay(s) & ~ ~ ~12.34±4.15            & ~ ~5.01±5.16 & 110.33±5.34  \\ 
\hline
Collisions       & 32.31±5.34                 & 64.31±6.67   & 47±6.78      \\
\hline
\end{tabular}
}
\end{table}

\noindent\textbf{(5) What are the strengths and limitations of the various CFLight extension methods? In which scenarios are these methods most appropriately applied?} CFLight-Loss leverages SCM-derived action advantages and jointly optimizes safety constraints and efficiency in a unified loss function, demonstrating flexibility but higher sensitivity to hyperparameters. CFLight-Q enhances interpretability by directly optimizing Q-values under CF scenarios, yielding near-zero collision rates, albeit at the cost of increased complexity and reliance on Q-network convergence. Each variant offers distinct strengths and trade-offs, making them suitable for different deployment priorities. We also create a table comparing the advantages and disadvantages of CFLight against other safe methods.

\noindent\textbf{(6) How do you identify which action/state in the trajectory was causal?} To identify causal actions/states, we compare outcomes by backtracking to pre-collision states and executing safe actions. If a collision is avoided, it reveals causal relationships. Our CTC generates tuples $(s, a_{cf}, r_{cf}, s_{cf})$ for capturing these causal relationships.

\noindent\textbf{(7) Why is it necessary to assume that $\mathcal{M}$ is monotonic in order to predict CF in an SCM?}  The monotonicity assumption on $\mathcal{M}_1$ and $\mathcal{M}_2$ with respect to the exogenous variables $U_s, U_r$ is essential for counterfactual prediction. Specifically, given observed data $(S,A,S',R)$ and the structural equations:
\begin{equation}
S' = \mathcal{M}_1(S,A,U_s), \quad R = \mathcal{M}_2(S,A,U_r),
\end{equation}
identifying the hidden variables $U_s$ and $U_r$ from the observations requires inverting $\mathcal{M}_1$ and $\mathcal{M}_2$. If $\mathcal{M}_1$ and $\mathcal{M}_2$ are monotonic in $U_s$ and $U_r$ respectively (for fixed $S,A$), then these mappings are invertible and the values of $U_s,U_r$ are uniquely determined:
\begin{equation}
U_s = \mathcal{M}_1^{-1}(S,A,S'), \quad U_r = \mathcal{M}_2^{-1}(S,A,R).
\end{equation}
This uniqueness is critical for generating consistent counterfactual outcomes under a hypothetical action $a_{cf}$, as it ensures that the same latent variables can be reused to simulate the counterfactual outcome:
\begin{equation}
S'_{cf} = \mathcal{M}_1(S,a_{cf},U_s), \quad R_{cf} = \mathcal{M}_2(S,a_{cf},U_r).
\end{equation}
Without monotonicity, the inversion may be non-unique or ill-defined, making counterfactual inference ambiguous.

\noindent\textbf{(8) How are CF actions selected to ensure safety for CFLight-Loss?} CF actions are selected by evaluating their advantage using the learned SCM and Q-function. Specifically: We monitor the real-world or simulated trajectory and identify actions leading to safety violations. For each detected unsafe state-action pair $(s, a)$, we generate a set of alternative candidate actions $\{a_{cf}^{(1)}, a_{cf}^{(2)}, \dots\}$ and simulate their outcomes using the SCM:
    \begin{equation}
        s'_{cf} = \mathcal{M}_1(s, a_{cf}, u_s), \quad r_{cf} = \mathcal{M}_2(s, a_{cf}, u_r)
    \end{equation}
For each candidate action, we compute its CF advantage:
    \begin{equation}
        A_{CF}(s, a_{cf}) = Q(s, a_{cf}) - V(s)
    \end{equation}
We choose the CF action $a_{cf}$ with the highest advantage, under the constraint that it does not trigger unsafe outcomes as predicted by the SCM (i.e., $r_{cf}$ indicates safety). This process balances \emph{efficiency} and \emph{safety}, enabling the agent to improve policy performance while avoiding unsafe behaviors.

\begin{figure}
\centering
\includegraphics[width=0.48\textwidth]{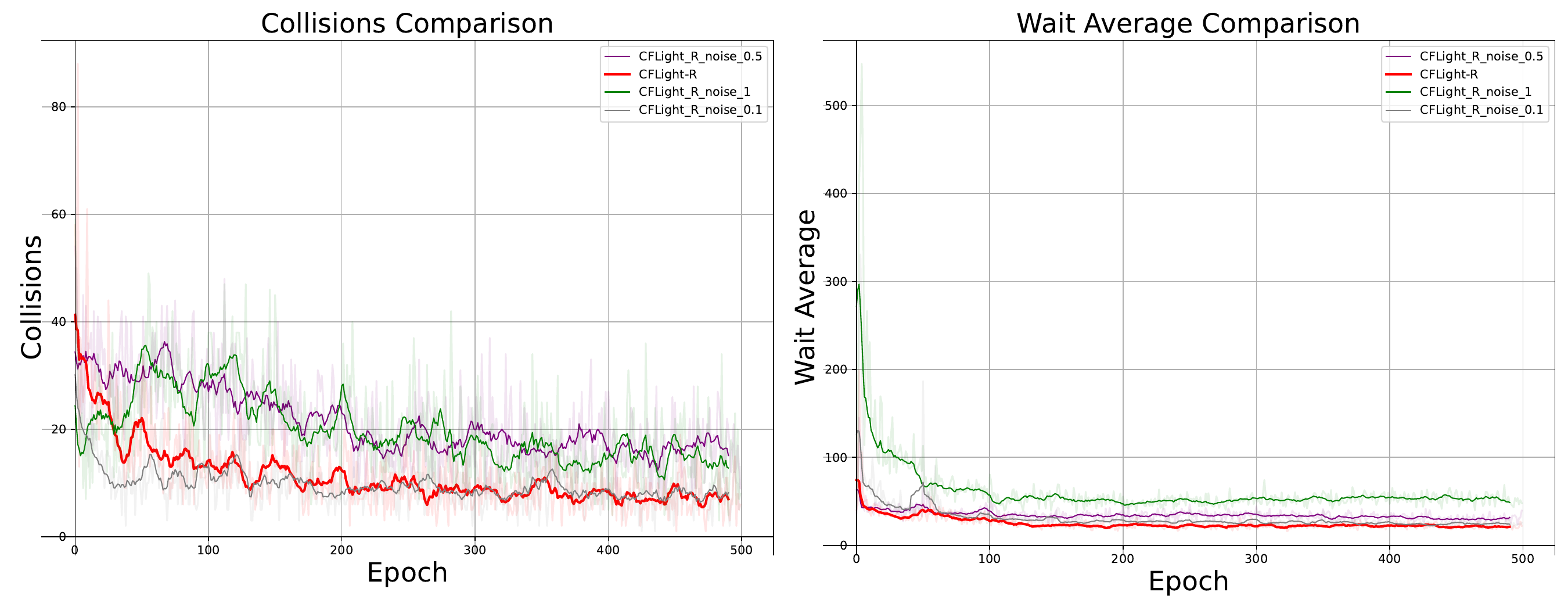}
\caption{Performance under noise of CFLight-R.}
\label{noise_compare}
\end{figure}
\noindent\textbf{(9) How sensitive is the method to inaccurate SCM predictions under varying traffic conditions or noise?} Figure~\ref{noise_compare} shows that, CFLight-R exhibits moderate sensitivity to inaccurate SCM predictions under varying traffic conditions, with noise modeled as $N(0, scale)$ and states normalized to the [0, 1] range. The Collisions Comparison shows that higher noise (0, 1) results in slightly more collisions and variability compared to lower noise (0, 0.5). Similarly, the Wait Average Comparison reveals a higher initial wait time and fluctuations with noise scale 1, compared to the smoother decline with scale 0.5, with the impact of noise diminishing over epochs. In the future, we will explore more control methods to improve the robustness of CFLight like~\cite{wu2023transformerlight,li2025fuzzylight, lirobustlight} and \textbf{more discussion and details as provided in Appendix~\ref{more_discu}}.

\section{Conclusion}

In this study, we introduced a flexible ``CF+X'' safe reinforcement learning framework, integrating counterfactual learning with the ``X'' module consisting of adjustable safety and efficient RL components. Within this framework, we developed a novel algorithm  called CFLight, addressing CF and safety concerns while smoothly integrating with existing RL methods for TSC. Our method was evaluated in the context of safe and efficient traffic management at intersections, showcasing an impressive 93.1\% reduction in collision rates compared to DQN while concurrently enhancing traffic efficiency. This work holds promise for broader applications in safe reinforcement learning, with potential implications for autonomous systems and robotics, offering a robust solution to enhance safety in complex decision-making scenarios.

\section*{Acknowledgments}
This work is supported by Fundamental and Interdisciplinary Disciplines Breakthrough Plan of the Ministry of Education of China (Grant No. JYB2025XDXM910), National Natural Science Foundation of China (Grant Nos. 42595590, 42595593),  Talent Scientific Fund of Lanzhou University (Grant No. 561120208), and Supercomputing Center of Lanzhou University. G. Yu contributed to this work in a strictly personal capacity as an independent researcher. This research was conceptualized, executed, and completed prior to the implementation of Australia’s 2025 policy updates regarding international collaborations. G. Yu did not receive any funding, compensation, in-kind assistance, or material support, directly or indirectly, from any grants (including the grants listed above) or from any other sources for this work. No institutional funds, resources, facilities, equipment, or work hours of his affiliation were utilized. This contribution falls outside the scope of his institutional employment and complies with applicable intellectual property regulations.

\clearpage

\bibliographystyle{IEEEtran}
\bibliography{sp}

\appendix

\section{Detailed Algorithm}
\label{detail_algo}
In this section we describe the other algorithms implemented by our CF+"X" framework and we describe the SCM training and inferece time. 

\subsection{SCM Training Detail}
In our experiments training the SCM, we utilized NVIDIA A100 GPUs to ensure robust computational performance. To stabilize the quality of data generation, we updated the generator and encoder five times for each discriminator update. To address gradient saturation issues, we assigned a fake label of 0.1 and a real label of 0.9, which provided more consistent gradient signals from the discriminator during early training phases, thereby enhancing the generator's learning efficiency. Furthermore, we conducted both training and inference experiments to assess the computational time impact of SCM. Using a buffer size of 20,000 and collision data ratios of 10\%, 20\%, 30\%, and 40\%, we trained the SCM model for 500 epochs. The corresponding training time is 17.92s and inference times are summarized in the Table~\ref{table:infe}.

\begin{table}[h]
\centering
\caption{Generation Performance for Different Collision Ratios}
\resizebox{0.8\linewidth}{!}{
\begin{tabular}{|c|c|c|}
\hline
\textbf{Collision Ratio (\%)} & \textbf{Generation Time (s)} & \textbf{Generated Data Quantity} \\
\hline
10.0 & 0.79 & 2000 \\
\hline
20.0 & 1.15 & 4000 \\
\hline
30.0 & 1.67 & 6000 \\
\hline
40.0 & 2.32 & 8000 \\
\hline
\end{tabular}
}
\label{table:infe}
\end{table}
It can be observed that the generation time increases with the rise in the amount of collision data. However, the time for a single data generation is 0.27 seconds, meeting real-time requirements.

\subsection{CFLight-Loss and CFLight-Q}
We introduce the other
four CFLight methods, as listed below:

\textbf{CFLight-Loss}: 
We extend the SafeLight method~\cite{du2023safelight} called CFLight-Loss. The safe component is integrated into the loss function, effectively transforming the learning process into a multi-task method. In multi-task learning, the model is trained to minimize a loss function that encompasses various objectives, each represented by a specific parameter vector, denoted as $\lambda$. 
These optimizing tasks may differ in their significance and priority~\cite{zamir2018taskonomy}.
To balance the different optimizing tasks, one adopts simple average or weighted sum techniques. 
By using these methods, the overall loss function $L(., ., \lambda)$ is represented as a sum of individual task losses, such as $L(., ., \lambda) = \sum_i \lambda_i * L^i(., .)$. In our case, we extend this principle and incorporate an additional optimizing task, specifically minimizing collisions, into the existing RL model. The resulting loss function is modified as follows:
\begin{equation}
    \label{safelightLoss}
    \begin{gathered}\mathcal{J}=\ c_1 \sum_s \mathcal{P}(s)\left[Q_{\text {target }}(s, a)-Q(s, a ; \theta)\right]^2+ 
 c_2 \Delta \\
  \Delta = \begin{cases}
\mathcal{D}_{\text{KL}} \left( A_{AC} \left( \pi_{H}(s) \right) \middle\| A_{AC}(s, a; \theta) \right), & \text{if Real World} \\
\mathcal{D}_{\text{KL}} \left( A_{\text{AC}} \left( \pi_{H}(s) \right) \middle\| A_{\text{CF}}(s, a_{\text{cf}}; \theta) \right), & \text{if Counterfactual World}
\end{cases}
    \end{gathered}
\end{equation}

By introducing the collision-minimization and CF task into the loss function, our RL model gains the ability to focus on both the original objectives and the safe aspect, ensuring a comprehensive learning process that accounts for multiple optimization objectives simultaneously, $c_1=0.5,c_2=0.5$.

\textbf{CFLight-Q}: We extend 
Synthetic Value (Syn-Q) \cite{gong2020multi} called CFLight-Q.
To obtain a
synthetic Q function, the linearly weighted sum of Q-values
for the three objectives is used, as shown in Equation \ref{synQ}.
\begin{center}
\begin{align}
    \label{synQ}
Q(s, a)=b_1\frac{Q_{Effe}(s, a)-Q_{Effe, \min }(s, a)}{Q_{Effe, \max }(s, a)-Q_{Effe, \min }(s, a)}+ \\ \notag
b_2 \frac{Safe_v(s, a)-Safe_{v, \min }(s, a)}{Safe_{v, \max }(s, a)-Safe_{v, \min }(s, a)} + \\\notag
b_{cf} \frac{Q_{cf}(s, a)-Q_{cf, \min }(s, a)}{Q_{cf, \max }(s, a)-Q_{cf, \min }(s, a)}
\end{align}
\end{center}

Where $Q(s,a)$ is the synthesized Q-function to control the generation of the next action, $Q_{Effe}$ determines the q value of the efficient component, $Safe_v$ and $Q_{cf}$ determines safety, they are normalized using max-min values, respectively, and the weights $b_1=0.5,b_2=0.5,b_{cf}=0.5$ denote importance.
Reward is divided into waiting time $r$, number of collisions $Safe_r$ and CF reward $r_{cf}$.

\begin{figure}[!htbp]
    \centering\includegraphics[width=0.48\textwidth]{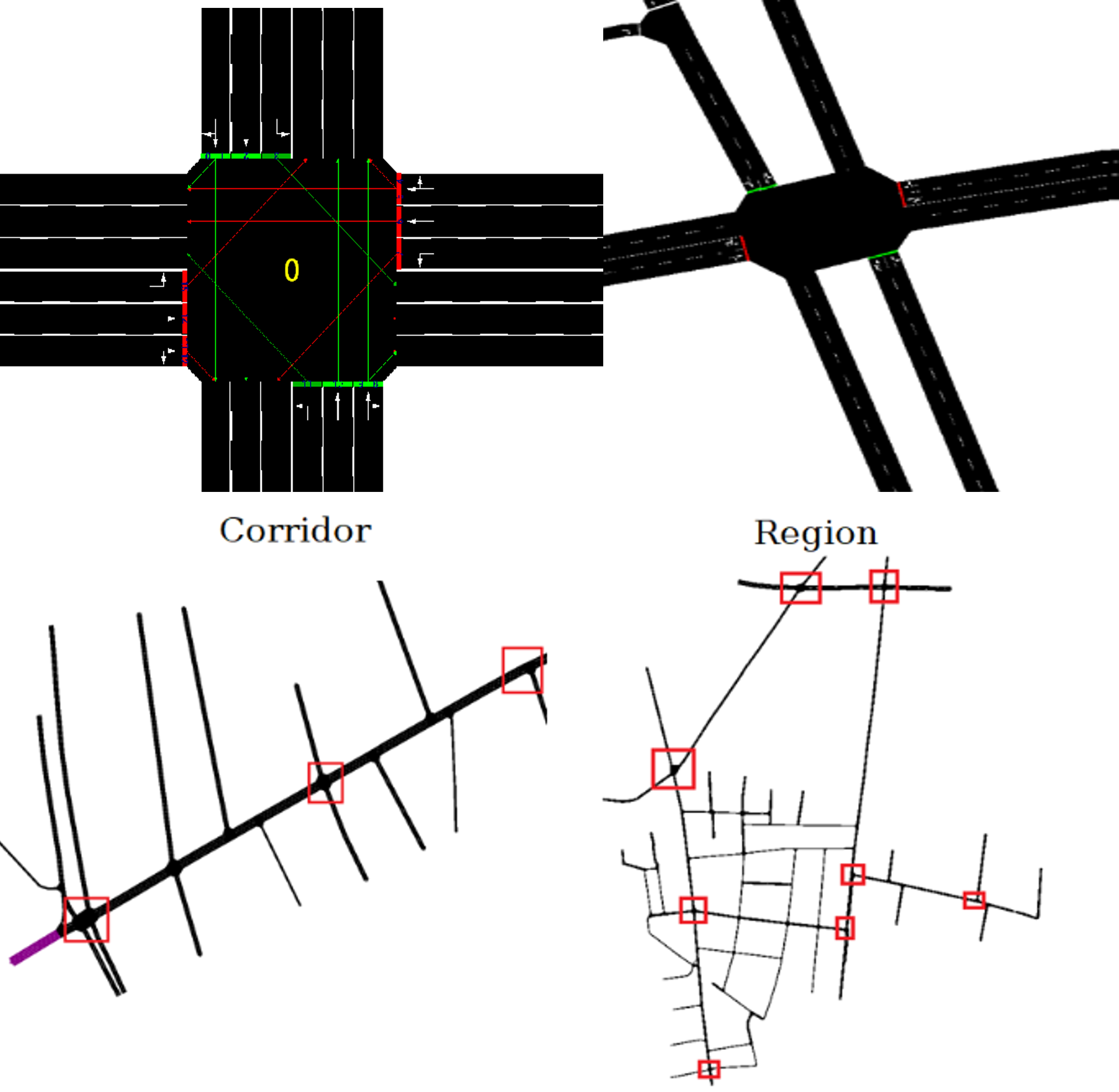}
    \caption{In the upper figure, the first one is the synthetic dataset, and the second one is Cologne 1. In the lower figure, the first one is Cologne 3, and the second one is Cologne 8.}
    \label{fig:dataset}
\end{figure}

\begin{table}
\renewcommand\arraystretch{1.2}
\centering
\caption{Network structures of SCM $\mathcal{M}$}
\label{table:network_structures}
\begin{tabular}{|c|c|c|} 
\hline
               & Hidden Layers & Neurons Per Layer                          \\ 
\hline
Generator~$\mathcal{M}_1$ & 3             & 600 $\rightarrow$ 600 $\rightarrow$ ~3600  \\ 
\hline
Generator~$\mathcal{M}_2$  & 3             & 600~$\rightarrow$ 600 $\rightarrow$ 2      \\ 
\hline
Encoder        & 3             & 600~$\rightarrow$~600~$\rightarrow$~7203   \\ 
\hline
Discriminator  & 3             & 600~$\rightarrow$~600~$\rightarrow$ 1        \\
\hline
\end{tabular}
\end{table}

\begin{table*}[t]
\centering
\small 
\setlength{\tabcolsep}{4pt} 
\caption{Comparison of CFLight-R, CFLight-Loss, and CFLight-Q Methods}
\label{table:compare}
\resizebox{0.9\linewidth}{!}{
\begin{tabular}{l|p{4cm}|p{4cm}|p{4cm}} 
\toprule
\textbf{Aspect} & \textbf{CFLight-R} & \textbf{CFLight-Loss} & \textbf{CFLight-Q} \\ 
\midrule
\textbf{Components} & $CF_r + Safe_r + Effe_r$ & $CF_l + Safe_l + Effe_r$ & $CF_q + Safe_r + Effe_r$ \\ 
\midrule
\multirow{3}{*}{\textbf{Advantages}} 
& High safety: fewer collisions than baselines. & Uses SCM for action advantages, less reliant on experts. & High interpretability via Q-value optimization. \\
& Balances safety and efficiency, Pareto optimal. & Single model optimizes dual objectives. & Q-value estimation enables automation vs. Safe-Act. \\
& Reward modification optimizes both objectives. & Flexible for safety constraints. & Near-zero collision rates. \\ 
\midrule
\multirow{3}{*}{\textbf{Disadvantages}} 
& Higher delay vs. non-CF methods. & Prioritizes safety, lower efficiency. & Relies on Q-network convergence. \\
& Complex due to counterfactual trajectories. & Sensitive to hyperparameters. & Efficiency trade-off in high-delay cases. \\
& & Limited experimental data. & Complex Q-network integration. \\
\bottomrule
\end{tabular}
}
\end{table*}

\section{Hyperparameter}
Detailed parameter settings are given in Table~\ref{table: hyperparameter}. It contains the SUMO simulator, the TSC hyperparameters and  the SCM model hyperparameters.

\begin{table}
\centering
\caption{Hyperparameters}
\label{table: hyperparameter}
\scalebox{0.8}{
\begin{tabular}{ccc} 
\toprule
Hyperparameter type                                                                              & Hyperparameter name           & Setting               \\ 
\hline
\multirow{10}{*}{\begin{tabular}[c]{@{}c@{}}SUMO~\\hyperparameter\end{tabular}}                  & jmIgnoreJunctionFoeProb       & 0.4                   \\
                                                                                                 & speedDev                      & 0.1                   \\
                                                                                                 & length                        & 4.3                   \\
                                                                                                 & minGap                        & 1                     \\
                                                                                                 & collision.mingap-factor       & 1                     \\
                                                                                                 & collision.check-junctions     & true                  \\
                                                                                                 & collision.action              & teleport              \\
                                                                                                 & ignore-accidents              & false                 \\
                                                                                                 & collision.stoptime            & 1                     \\
                                                                                                 & step-length                   & 0.5                   \\ 
\hline
\multirow{10}{*}{\begin{tabular}[c]{@{}c@{}}GAN \\hyperparameter\end{tabular}}                   & generator\_learning\_rate     & 0.0004                \\
                                                                                                 & encoder\_learning\_rate       & 0.0004                \\
                                                                                                 & discriminator\_learning\_rate & 0.003                 \\
                                                                                                 & optimizer                     & AdamW                 \\
                                                                                                 & batch\_size                   & 128                   \\
                                                                                                 & epochs                        & 500                   \\
                                                                                                 & loss function                 & BCE,MSE,CrossEntropy  \\
                                                                                                 & $\lambda$                     & 1                     \\
                                                                                                 & $\beta$                       & 1                     \\
                                                                                                 & $u$ dim                       & 4                     \\ 
\hline
\multirow{12}{*}{\begin{tabular}[c]{@{}c@{}}TSC RL agent \\training hyperparameter\end{tabular}} & discount($\gamma$)            & 0.99                  \\
                                                                                                 & learning rate                 & 0.0003                \\
                                                                                                 & action dim                    & 10                    \\
                                                                                                 & target critic($\tau$)         & 0.001                 \\
                                                                                                 & buffer capacity               & 12000                 \\
                                                                                                 & epochs                        & 800                   \\
                                                                                                 & batch\_size                   & 128                   \\
                                                                                                 & learning\_rate                & 0.001                 \\
                                                                                                 & target update time            & 4                     \\
                                                                                                 & hidden size                   & 64                    \\
                                                                                                 & loss function                 & MSE                   \\
                                                                                                 & optimizer                     & Adam                  \\
                                                                                                 & $w_1,w_2$                     & 0.5,0.5             \\

                                                                                                 & trainGap                     & 50  \\     
\bottomrule
\end{tabular}
}
\end{table}

\section{More Experiments Detail}
\label{more_exp_detail}
In this section, we describe the network structure, datasets and compared methods.
\subsection{Network Structure}
We describe the network structures of the TSC network and the SCM network. The TSC network uses the same architecture as SafeLight and IPPO. The SCM network employs linear hidden layers with ReLU activation functions and uses a Sigmoid activation function for the final layer of the Discriminator, detailed in Table~\ref{table:network_structures}.

\subsection{Datasets}
We experiment with a synthetic intersection~\cite{liang2019deep} and a real-world intersection~\cite{mei2022libsignal}.

\textbf{Synthetic:} The synthetic intersection comprises four methods and twelve one-way vehicular movements with synthetic traffic flow adopted into SUMO and Figure~\ref{fig:dataset}).

\textbf{Real-world:} The real-world dataset is derived from the Cologne, Germany intersection, incorporating real traffic patterns into the SUMO environment. This intersection features 8 approaching lanes. In both cases, the SUMO simulation environment assumes accurate vehicle detection, consistent deceleration rates (modifiable as needed), and permitted left turns within the timing plan. 

\section{More Discussion}
\label{more_discu}

\noindent\textbf{(1) Why are compositional Q-values necessary if CFLight-R already performs well?} CFLight-Q enhances interpretability by optimizing Q-values with CF scenarios, while CFLight-R relies on reward shaping. 

\noindent\textbf{(2) Do there have theoretical analysis of CFLight-Loss?} The efficacy of CFLight-Loss is grounded in the convergence of the Q-learning algorithm to the optimal action-value function $Q^*(s,a)$, as discussed in the Bellman-based formulation. The distributional distance between the action taken in the real world and the counterfactual safe alternative is captured by:
\begin{equation}
\label{CF-action}
CF_{l} = D_{KL}(A_{AC}(\pi_H(s)) \,\|\, A_{CF}(s, a_{cf}; \theta)),
\end{equation}
where $A_{AC}(\pi_H(s))$ denotes the advantage distribution associated with real-world unsafe actions (e.g., collisions), and $A_{CF}(s, a_{cf}; \theta) = Q(s, a_{cf}; \theta) - V(s)$ represents the counterfactual advantage distribution.

Since $Q(s,a;\theta)$ is obtained through temporal difference updates based on the Bellman equation, its convergence to $Q^*(s,a)$ is guaranteed under standard Q-learning conditions:
\begin{enumerate}
    \item The state and action spaces are discrete and finite, as the traffic state is quantized into threshold-based bins.
    \item All relevant $(s,a)$ pairs, including those arising from real and counterfactual evaluations, are sufficiently sampled.
    \item The learning rate $\alpha$ decays appropriately, and the discount factor $\gamma$ remains within $(0,1)$.
\end{enumerate}

Therefore, as the underlying $Q(s,a;\theta)$ converges to $Q^*(s,a)$, the CF advantage $A_{CF}$ becomes a reliable estimate of potential safety under alternative actions. The KL divergence $D_{KL}(\cdot \,\|\, \cdot)$ between $A_{AC}$ and $A_{CF}$ thus quantifies the safety gap and is grounded in convergent Q-values.

\noindent\textbf{(3) Since training uses a simulation, could CF trajectories be generated by querying the traffic simulator directly, avoiding SCM learning?} While it is indeed possible to generate CF trajectories directly by querying the traffic simulator during training, our use of the SCM serves a different purpose. The SCM is designed to approximate the dynamics of the real world, enabling CF inference under real-world constraints where direct access to a simulator is not feasible. In other words, the SCM allows us to predict how the system would have evolved under alternative actions in the real world, beyond the simulation environment.

\end{document}